\documentclass[sn-mathphys,Numbered]{sn-jnl}

\usepackage{adjustbox}

\usepackage{graphicx}%
\usepackage{multirow}%
\usepackage{amsmath,amssymb,amsfonts}%
\usepackage{amsthm}%
\usepackage{mathrsfs}%
\usepackage[title]{appendix}%
\usepackage{xcolor}%
\usepackage{textcomp}%
\usepackage{manyfoot}%
\usepackage{booktabs}%
\usepackage{algorithm}%
\usepackage{algorithmicx}%
\usepackage{algpseudocode}%
\usepackage{listings}%
\usepackage{rotating}%
\usepackage{threeparttable}%
\usepackage{multicol}%
\usepackage{tabularx}%
\usepackage{subcaption}%

\theoremstyle{thmstyleone}%
\theoremstyle{thmstyletwo}%

\theoremstyle{thmstylethree}%

\begin{document}

\title[Innovation and Word Usage Patterns in Machine Learning]{Innovation and Word Usage Patterns in Machine Learning}


\author[1,2,3]{\fnm{V\'itor} \sur{Bandeira Borges}}\email{vitorbborges@gmail.com}

\author*[1,2,4]{\fnm{Daniel} \sur{Oliveira Cajueiro}}\email{danielcajueiro@unb.br}
\equalcont{These authors contributed equally to this work.}

\affil*[1]{\orgdiv{Department of Economics}, \orgname{University of Brasilia}, \orgaddress{\street{Campus Universitario Darcy Ribeiro - FACE}, \city{Brasilia}, \postcode{70910-900}, \state{DF}, \country{Brazil}}}

\affil[2]{\orgdiv{Machine Learning Laboratory in Finance and Organizations (LAMFO)}, \orgname{University of Brasilia}, \orgaddress{\street{Campus Universitario Darcy Ribeiro - FACE}, \city{Brasilia}, \postcode{70910-900}, \state{DF}, \country{Brazil}}}

\affil[3]{\orgdiv{Tutorial Education Program (PET)}, \orgname{University of Brasilia}, \orgaddress{\street{Campus Universitario Darcy Ribeiro - FACE}, \city{Brasilia}, \postcode{70910-900}, \state{DF}, \country{Brazil}}}

\affil[4]{\orgdiv{National Institute of Science and Technology for Complex Systems (INCT-SC)}, \orgname{University of Brasilia}, \orgaddress{\street{Campus Universitario Darcy Ribeiro - FACE}, \city{Brasilia}, \postcode{70910-900}, \state{DF}, \country{Brazil}}}


\abstract{In this study, we investigate the dynamic landscape of machine learning research evolution. Initially, through the use of Latent Dirichlet Allocation, we determine pivotal themes and fundamental concepts that have emerged within the realm of machine learning. Subsequently, we undertake a comprehensive analysis to track the evolutionary trajectories of these identified themes.  To quantify the novelty and divergence of research contributions, we use the Kullback-Leibler Divergence metric. This statistical measure serves as a proxy for ``surprise", indicating the extent of differentiation between the content of academic papers and the subsequent developments in research. We also analyze the roles of prominent researchers and the significance of academic venues (journals and conferences) in the field of machine learning.
}




\maketitle

\section{Introduction}\label{Introduction}

Machine learning has become ubiquitous in many fields today, ranging from economics, finance, medicine, and healthcare to marketing and transportation. It enables businesses and organizations to extract valuable insights from large amounts of data and make predictions based on patterns and trends that would otherwise be difficult or impossible to detect.  By extracting valuable information from data and automating several tasks, machine learning can save time and reduce costs while improving accuracy and efficiency. 

This paper explores the evolution of research on machine learning, using a large collection of papers in the field. We begin by identifying the main topics and dominant concepts within each area. Next, we trace the emergence of these key concepts and analyze how they have evolved over time. Finally, we examine the relationships between the current state-of-the-art and past developments, providing insight into the ways in which the field has progressed. In addition, we investigate the importance of the main machine learning venues and the more frequent and prominent authors in disseminating the theme. 

A fundamental building block of our work is the Latent Dirichlet Allocation (LDA) model \citep{Blei2003}. We use it to identify the main topics and concepts of the field of machine learning. Associated with the application of this technique we also use a coherence metric \citep{roder2015} to indicate the suitable number of the topics of the model. With the papers divided by topics, we are able to analyze the trends of the machine learning field. In addition, we use the Kullback-Leibler Divergence (KLD) as the notion of ``surprise" that measures the statistical divergence between the contributions of papers in subsequent events. We can use surprise to measure the divergence of one piece of news from previous ones called ``Novelty" and the divergence of one piece of news from the later ones called ``Transience". With these concepts in hand, we are able to define the concept of ``Resonance" as the difference between Novelty and Transience \citep{BarronFRev}. Thus, we may evaluate the role of the most relevant venues used to disseminate the knowledge of machine learning and the role of the most frequent and prominent researchers.

Our data comes from 25 very relevant machine learning venues that include annals of conferences and prestigious journals. The method we use to choose these venues is based on a two step procedure. In the first step, we look into a series of popular machine learning blogs for the most popular indications. In the second step, we check if these indications belong to the list of the top venues based on the Google scholar h5-index considering the subcategories of ``Artificial Intelligence", ``Computational Linguistics", ``Data Mining and Analysis" and ``Engineering and Computer Science".  

Our work relates to other studies that tell the history and the evolution of the machine learning field such as \citet{langley2011changing} and \citet{fradkov2020early}. It also naturally relates to the work of \citet{BarronFRev} that uses the concepts of novelty, transience and resonance to study how ideas are created, ignored or propagated in the context of the French revolution. In this same context, we may also cite \citet{jurafsky2008} that also applied LDA to to the ACL Anthology to analyze historical trends in the
field of Computational Linguistics. In addition, \citet{savov2020identifying} proposes a a LDA based method to estimate a innovation score of a given paper. 

Our work is organized as follows. Section \ref{sec:methods} describes the procedures we adopt to tune and estimate the models. We detail the data set we use in Section \ref{sec:data} and present the results in Section \ref{sec:results}.  Section \ref{sec:conclusion} summarizes and concludes the work.

\section{Methods}\label{sec:methods}

We have divided this section into three segments. In Section \ref{lda}, we provide an overview of the LDA model and the methodology for determining the optimal number of topics. Moving on to Section \ref{sec:modellingTrends}, we demonstrate the application of the LDA-derived topics in monitoring the progression of machine learning research. Further in Section \ref{KLD}, we revisit the KLD metric, elucidating its utility in delineating the constructs of novelty, transience, and resonance.

\subsection{Latent Dirichlet Allocation}
\label{lda}

The LDA model, introduced by \cite{Blei2003}, is a generative probabilistic model for topic modeling.  It is based on the assumption that documents are mixtures of topics, and each topic is a distribution over words. In order to be mathematically precise, we use the following notation. The vocabulary $\mathcal{V} = \{ w_1, \ldots, w_i, \ldots, w_{N_{V}} \}$ is the set of all distinct words (present in all documents) and $I_{V} = \{ 1, \ldots, N_{V} \}$ is the set of all word indexes, where  $N_V$ is the number of distinct terms, i.e., the size of the vocabulary.  A document $d = [w_{i_1}, \ldots, w_{i_k}, \ldots, w_{i_{N_d}}]$ consist of a list of $N_d$ non-unique consecutive words ($1 \leq k \leq N_d$ and $i_k \in I_{V}$), while $\mathcal{V}^{d}$ is the vocabulary that appears in the document $d$. 
Based on the assumption that the number of topics $K$ is fixed and known, LDA assumes the following generative process for each document $d$ in a corpus $D$:

\begin{enumerate}
    \item Choose $N_d\sim \text{Poisson}(\xi)$;
    \item Choose $\theta\sim \text{Dir}(\alpha)$, where the parameter $\alpha$ is $K$ vector of positive components that we need to estimate;
    \item For each of the $N_d$ words of $w_n$;
    \begin{enumerate}
        \item Choose topic $z_n\sim \text{Multinomial}(\theta)$;
        \item Choose word $w_n$ from $p(w_n|z_n,\beta)$, a multinomial probability conditioned to topic $z_n$, where the word probabilities are parametrized by a $K\times N_V$ matrix $\beta=p(w^j=1|z^i=1)$ for all $j\in \{1,...,N_V\}$ and all $k \in \{1,...,K\}$.
    \end{enumerate}
    
\end{enumerate}

    In the first step, LDA chooses the number of words in the document, denoted by $N_d$, from a Poisson distribution with parameter $\xi$. This determines the length of the document. Then, it chooses the document's topic proportions, denoted by $\theta$, from a Dirichlet distribution with parameter $\alpha$. This step determines the distribution of topics within the document. Thus, for each of the $N_d$ words in the document, it chooses the word's topic, denoted by $z_n$, from a Multinomial distribution with probabilities determined by the document's topic proportions $\theta$, which determines which topic the word belongs to. And, finally, it chooses the specific word, denoted by $w_n$, from a Multinomial distribution conditioned on the chosen topic $z_n$, where, as above-mentioned, the word probabilities are parameterized by a $K \times N_V$ matrix $\beta$, where $N_V$ is the vocabulary size. 
    This generative process is repeated for each document in the collection. There are different ways to estimate this model. In our paper, we estimate it using the online variational Bayes algorithm due to \cite{hoffman2010online}\footnote{We use the implementation available in the Gensim Python library \citep{rehurek_lrec}.}.

Selecting an appropriate number of topics stands as a fundamental prerequisite for executing the LDA model. Notably, it is imperative to acknowledge that evaluating the efficacy of LDA, akin to other unsupervised models, presents challenges stemming from the absence of labels that can serve as benchmarks to validate the accuracy of outcomes. While the most effective approach to appraising unsupervised models involves human assessments, such an evaluation methodology can incur substantial costs and, in cases of extensive datasets, may even become unfeasible.
Consequently, within this contextual framework, a prevalent recourse involves the utilization of metrics that capture the frequency of co-occurrences within a given corpus. These metrics find application within the domain of LDA, hinging upon the identification of the words per topic and the analysis of their co-occurrences within the corpus.
In our paper,  we use the Normalized Pointwise Mutual Information (NPMI) \citep{bouma2009normalized} coherence measure . Let the Pointwise Mutual Information (PMI) be given by \citep{church1990word}

\begin{equation}
        PMI(w_i,w_j) = log \frac{P(w_i,w_j) + \epsilon}{P(w_i) \cdot P(w_j)} \\
\end{equation}
\noindent where $P(w_i, w_j)$ is the joint probability of words $w_i$ and $w_j$ measured in a fixed-size window in the text,  $P(w_i)$ and $P(w_j)$ are the individual probabilities of the words and $\epsilon$ is a small number added to the joint probability to avoid logarithm of zero. Thus, PMI is a measure of how much the actual probability of a particular co-occurrence of words  $p(w_i,w_j)$ differs from what we would expect it to be on the basis of the probabilities of the individual words and the assumption of independence $p(w_i)p(w_j)$.

The NPMI is a normalized form of the PMI measure. Although there are different ways to normalize the PMI, \citet{bouma2009normalized} normalizes it by the $(-log(P(w_i,w_j) + \epsilon))$, since this option normalizes both the upper and the lower bound. Thus, we may write the NPMI by 

\begin{equation}
    NPMI(w_i,w_j) = \left( \frac{PMI(w_i,w_j)}{-log(P(w_i,w_j) +\epsilon)} \right).
\end{equation}

In order to  quantify how semantically related the words within a topic are, we may evaluate the coherence of a topic $T_k$ using 

\begin{equation}
    C_V(T_k) = \frac{2}{|T_k|(|T_k|-1)} \sum_{i=1}^{|T_k|-1} \sum_{j=i+1}^{|T_k|} NPMI(w_i, w_j),\\
\end{equation}

where $|T_k|$ is the number of words in the topic $T_k$.

Aiming at considering the quality of all the topics together, we average $C_V$ to get 

\begin{equation}
    \overline{C_V} = \frac{1}{K} \sum_{k=1}^{K} C_V(T_k),
\end{equation}
\noindent where $K$ is the number of topics. An important characteristic of this coherence measure is its high correlation with human judgment in assessing the quality of topics \citep{roder2015}.

\subsection{Modelling trends}
\label{sec:modellingTrends}

As in \cite{jurafsky2008}, in order to capture the temporal dynamics among topics, we evaluate the observed probability of each topic within specific time intervals. This probability assessment involves calculating the average likelihood of each topic across the papers published during that period. This process is repeated for all topics across all distinct time periods under consideration.

\subsection{Kullback Leibler Divergence}
\label{KLD}

The Kullback-Leibler Divergence (KLD) \citep{kullback51}, also known as relative entropy, is a measure of information loss when an {\it observed} probability distribution $p$ is estimated using a {\it theoretical} distribution $q$. If the observed and theoretical distributions are the same ones, the divergence is zero. On the other hand, if we consider two vastly different  distributions, the divergence is very high, meaning a great loss of information due to misspecification. 

In the context of topic modeling, we can use KLD to quantify the dissimilarity between a document's topic distribution and a reference topic distribution.
 From an information retrieval perspective, we may interpret relative entropy as a measure of ``surprise'' when one document is expected and another is observed \citep{BarronFRev}. Given an LDA-generated set of probability distributions $p^{(j)} = (p_1^{(j)},p_2^{(j)},\ldots,p_K^{(j)})$, where $j$ indexes chronological order and $K$ is the number of topics, we may evaluate the {\it surprise} between times $j$ and $i$ as
\begin{align}
    \text{KLD }(p^{(j)}|p^{(i)})&=\sum_{k=1}^{K} p_k^{(j)} \log_2 \frac{p_k^{(j)}}{p_k^{(i)}},
\end{align}
\noindent where $K$, as before, is the number of topics\footnote{It is worth noting that, unlike the paper by \citet{BarronFRev}, our approach does not adhere strictly to a chronological ordering of the papers. Instead, we arrange the papers in chronological order by month, which serves as the temporal unit allowing us to reconstruct the paper sequence based on the publication dates provided by the academic venues. In addition, to assess the novelty and transience (that we will define below), we compute these measures for each paper relative to all papers published in the preceding period and calculate the average value.}.

We may define the {\it novelty} $\mathcal{N}_w(j)$ of the $j$-th document by the average surprise between itself and the past documents that took place in a time scale $w$:
\begin{align}
    \mathcal{N}_w(j)&=\frac{1}{w}\sum_{d=1}^{w} \text{ KLD }(p^{(j)}|p^{(j-d)}).
\end{align}

On the other hand, we may define the {\it transience} $\mathcal{T}_w(j)$ of the $j$-th document by the average surprise between itself and the future documents that will take place in a time scale $w$:
\begin{align}
\mathcal{T}_w(j) &= \frac{1}{w}\sum_{d=1}^{w} \text{ KLD }(p^{(j)}|p^{(j+d)}).    
\end{align}

We measure {\it resonance} $\mathcal{R}_w(j)$ as the difference between novelty and transience: 
\begin{align}
    \mathcal{R}_w(j) &= \mathcal{N}_w(j) - \mathcal{T}_w(j). 
\end{align}

We may interpret the resonance of a document in a corpus of news stories as an indicator of a novel subject that is capable of influencing the general direction of outlets, being written about again in the future. 

In addition, we may measure the expected resonance of any document given some
level of novelty with a linear model

\begin{equation}E[\mathcal{R}|\mathcal{N}]=\beta_{\mathrm{int}}+\beta_{\mathcal{N}}\mathcal{N}\label{eq:ressonancexinnovation}\end{equation}

\noindent and, using this linear equation, we may define {\it novelty effectiveness} $\Gamma$ as the rate at which resonance increases with novelty:

\begin{align}
\label{eq:novelty_effectiveness}
    \Gamma = \frac{\partial E[\mathcal{R}|\mathcal{N}]}{\partial \mathcal{N}} = \beta_{\mathcal{N}}.
\end{align}

Novelty effectiveness provides a nuanced understanding of the dynamics of speech influence. It highlights the delicate balance speakers must strike between novelty and resonance, and the inherent risk and reward associated with introducing novel ideas.

The time period parameter for calculating the average innovation between papers, denoted as $w$, was set to be equal to 12 months.  

\section{Dataset}\label{sec:data}

Our dataset consists of 25 venues related to machine learning, including both conference proceedings and periodicals. We choose these venues using a two step procedure. In the first step, we look into a series of popular machine learning sources for the most popular indications presented in Appendix \ref{app:urls}. In the second step, we check if these indications belong to the list of the top 20 publications based on the Google Scholar h5-index\footnote{This Google scholar tool is available at \href{https://scholar.google.com/citations?view_op=top_venues}{https://scholar.google.com/citations?view\_op=top\_venues}.} considering the subcategories of ``Artificial Intelligence", ``Computational Linguistics", ``Data Mining and Analysis" and ``Engineering and Computer Science". Of these, 24 venues were indexed on the Web of Science (WoS) database. The International Conference on Learning Representations, however, was not accessible in WoS, necessitating manual extraction from the \cite{dblp} API.

The dataset, comprising 168,757 publications, serves as the foundation for this research, which aims to scrutinize abstracts and their interrelationships throughout the field's history. The dataset includes 95,626 (56.66\%) papers in academic periodicals, 72,188 (42.78\%) conference papers, 940 publication series, and 3 books.
Among these papers, we are not able to use 4,001 of them because they do not have abstracts. We also exclude from our study 3,750 publications that we are not able to recover the date. It is worth mentioning that in most cases, the date of the publication is directly  available in the data extracted from the WoS. However, in some specific cases we have to directly deal with it. In particular, in some conferences, the date of publication was not available and we replaced this missing value by the date that the conference took place. In a small number of cases, the description of the date of the conference was given by the station of the year, namely {\it Summer}, {\it Autumn}, {\it Winter} and {\it Spring}. In these cases, we carefully looked into the correct date of the publication and replaced this information by it. Due to the lack of precision of this piece of data, we adopt the monthly granularity for our time series.

\begin{figure}[H]
  \centering
  \caption{Number of Publications per Year.}
  \includegraphics[width=\linewidth]{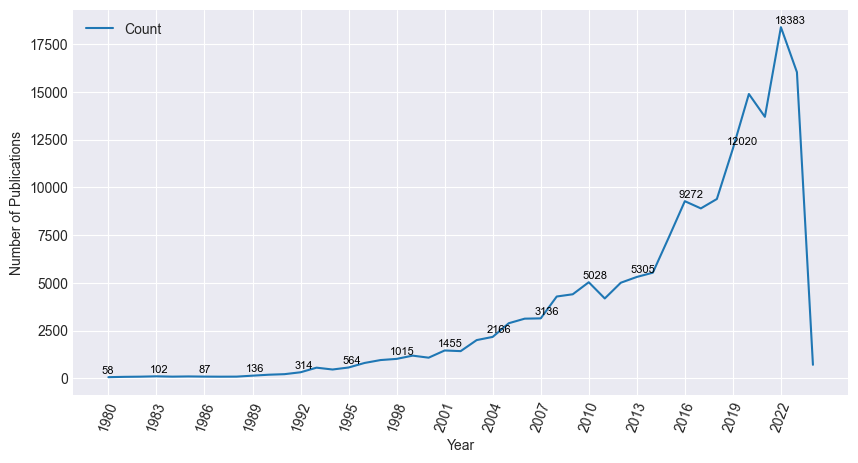}
  \label{fig:publications_yearly}
\end{figure}

The exponential growth of publications in our database is illustrated in Figure \ref{fig:publications_yearly}. This remarkable trend mirrors the recent upsurge of interest and resources devoted to machine learning \citep{jordan2015machine}, which has facilitated the swift generation and obsolescence of innovative concepts. The expansion of publications in both number and velocity is indicative of the dynamic nature of this field, where novel findings and ideas emerge at a rapid pace.

Table \ref{table:dataset_summary} offers a summary of the dataset, detailing the number of papers, accessible date ranges, and predominant publication types by venue. The complete dataset and associated code can be accessed through this paper's \href{https://zenodo.org/record/8298911}{Zenodo}.

\begin{sidewaystable*}[]
\centering
\caption{Summary of venues, counts, date ranges, and publication types in the dataset}
\label{table:dataset_summary}
\resizebox{\linewidth}{!}{%
\begin{tabular}{@{}l|l|l|l@{}}
Venue                                                                      & Count  & Date Range           & Primary Publication Type \\
\toprule
Neurocomputing                                                             & 17830  & Feb-1992 to Jan-2023 & Journal                        \\
Expert Systems with Applications                                           & 17273  & Dec-1992 to Apr-2023 & Journal                        \\
International Conference on Machine   Learning                             & 15785  & Oct-1997 to Dec-2022 & Conference                     \\
International Conference on Computer   Vision (ICCV)                       & 12051  & Jun-1995 to Feb-2022 & Conference                     \\
AAAI Conference on Artificial   Intelligence                               & 9246   & Nov-2008 to Feb-2021 & Conference                     \\
Applied Soft Computing                                                     & 8369   & Feb-2004 to Nov-2022 & Journal                        \\
Neural Computing and Applications                                          & 8296   & Sep-1997 to Dec-2022 & Journal                        \\
IEEE Transactions on Pattern Analysis and   Machine Intelligence           & 7226   & Jan-1986 to Dec-2022 & Journal                        \\
Knowledge-Based Systems                                                    & 6767   & Mar-1991 to Jan-2023 & Journal                        \\
Neural Information Processing Systems                                      & 6020   & Dec-1992 to Dec-2020 & Conference                     \\
Computer Vision and Pattern Recognition   Conference                       & 5786   & Jun-2000 to Jun-2019 & Conference                     \\
Meeting of the Association for   Computational Linguistics (ACL)           & 5771   & Jun-1993 to May-2022 & Conference                     \\
International Joint Conferences on   Artificial Intelligence Organization  & 5291   & Jul-2005 to Sep-2020 & Conference                     \\
Neural Networks                                                            & 5078   & Feb-1990 to Jan-2023 & Journal                        \\
IEEE Transactions on Neural Networks and   Learning Systems                & 5059   & Jan-2012 to Dec-2022 & Journal                        \\
Applied Intelligence                                                       & 4706   & Feb-1993 to Dec-2022 & Journal                        \\
Engineering Applications of Artificial   Intelligence                      & 4515   & Jan-1992 to Jan-2023 & Journal                        \\
International Conference on Learning   Representations                     & 4353   & May-2013 to Apr-2022 & Conference                     \\
European Conference on Computer Vision                                     & 4124   & Sep-1997 to Mar-2020 & Conference                     \\
IEEE Transactions on Fuzzy Systems                                         & 3766   & Feb-1994 to Dec-2022 & Journal                        \\
Journal of Machine Learning Research                                       & 2657   & Oct-2000 to Dec-2015 & Journal                        \\
International Conference on Artificial   Intelligence and Statistics       & 2526   & Apr-2014 to Mar-2022 & Conference                     \\
Conference on Empirical Methods in   Natural Language Processing           & 2479   & Feb-1999 to Nov-2021 & Conference                     \\
IEEE Transactions on Systems, Man, and   Cybernetics, Part B & 2104   & Feb-1996 to Dec-2012 & Journal                        \\
Artificial Intelligence Review                                             & 1679   & Feb-1993 to Dec-2022 & Journal                        \\
\bottomrule
Total                                                                      & 168757 & Jan-1986 to Apr-2023 & Journal                       
\end{tabular}
}
\end{sidewaystable*}

\section{Results}\label{sec:results}

AIn this section, we present our results. In Subsection \ref{result-topic_trends}, we present the discovered topic trends uncovered during our study. In Subsection \ref{results-ntr}, we delve into the assessment of novelty, transience, and resonance as key characteristics of machine learning research. Here, we examine the roles of authors and venues in shaping this field, evaluating their impact and influence. We present the details of our LDA implementation in Appendix \ref{app:ldaDetails}.

\subsection{Topics Trends}
\label{result-topic_trends}

The dynamics of scientific progress and the elements influencing the ascent and descent of academic interest in diverse subjects have been extensively debated among historians, sociologists, philosophers of science, and scientists themselves \cite{griffiths2004}. By reducing a corpus of scientific documents to a set of topics, we can enhance our understanding of the development of scientific pursuits and the driving forces behind these shifts.

In the following subsections, we utilize LDA and observed probability of each topic to extract the trends of specific relevant topics in the field of machine learning, including deep learning (Section \ref{deep_learning}), computer vision (Section \ref{computer_vision}), natural language processing (Section \ref{naturalLanguageProcessing}), reinforcement learning (Section \ref{reinforcementLearning}), and expert systems (Section \ref{expertSystems}). We conclude this section with Section \ref{historicalContexts}, which examines the potential impact of certain real-world events on machine learning research.

\subsubsection{Deep Learning}
\label{deep_learning}

As a prominent subfield of machine learning, deep learning focuses on the design and application of artificial neural networks, particularly those with multiple hidden layers, to address complex computational problems. Influential researchers in deep learning, such as \citet{Hinton2006}, \citet{LeCun1989}, and  \citet{Bengio2007}, have been instrumental in the development of the field.

This powerful approach has significantly propelled advancements in various areas of machine learning, such as computer vision \citep{Krizhevsky2012}, natural language processing \citep{Sutskever2014}, and speech recognition \citep{Hinton2012}, by enabling the extraction of hierarchical features and promoting the development of end-to-end learning systems.

\begin{figure}[H]
\centering
\caption{Deep Learning Related Topics}
\includegraphics[width=\linewidth]{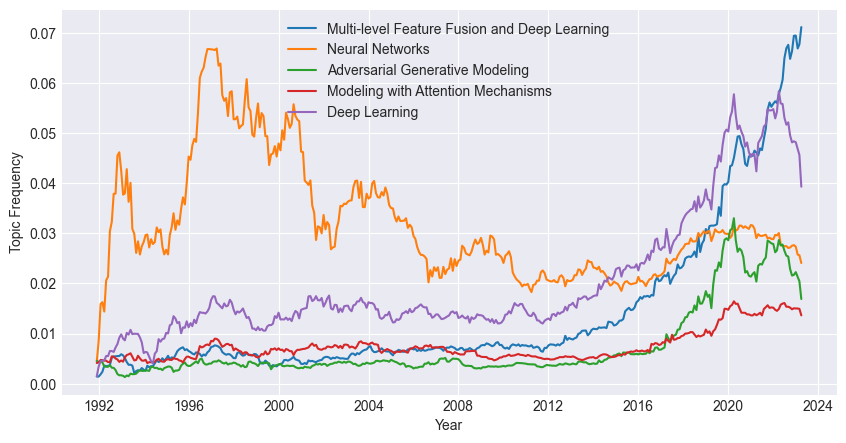}
\label{fig:deep_learning}
\end{figure}

Figure \ref{fig:deep_learning} illustrates the frequency of Deep Learning related topics over the past 30 years, clearly demonstrating the substantial evolution of the subfield within the last five to eight years. This figure exemplifies \cite{Kuhn1962} model of scientific evolution, which posits that a community's adoption of a new paradigm triggers a shift in focus, provoking debates and promoting advancements in novel areas. In the subfield's literature, the prevailing paradigm transitioned from \textit{Neural Networks} — a topic that primarily emphasized data representations and the design of network architectures for capturing features within data — to subjects that focus on improving model performance through novel training techniques, optimization algorithms, and architectural innovations.

Comparatively recent approaches, such as \textit{Adversarial Generative Modeling} (AGM) \citep{Goodfellow2014} and \textit{Modeling with Attention Mechanisms} \citep{bahdanau2014neural}, experienced an upsurge in a more condensed timeframe than \textit{Deep Learning}. The foundational work of Hinton, LeCun, and Bengio influenced the surge of publications in \textit{Multi-level Feature Fusion and Deep Learning}, which subsequently led to the emergence of AGM and Attention. These subjects encompass various aspects of model training, such as discovering better optimization methods, understanding the benefits of depth in neural networks, regularization techniques, and novel architectures that facilitate learning in complex domains.

\subsubsection{Computer Vision}
\label{computer_vision}

Computer vision, a multidisciplinary subfield of machine learning, focuses on enabling machines to interpret and comprehend visual information from their surroundings. Drawing on techniques from image processing \citep{Canny1986}, pattern recognition \citep{Duda2000}, and statistical learning \citep{Hastie2009}, it plays a crucial role in artificial intelligence by empowering systems to interact with and make sense of the visual world, facilitating applications in robotics, surveillance, healthcare, and autonomous vehicles.

During the 1990s, computer vision techniques were primarily based on rigorous mathematical analysis and quantitative aspects. Examples of models from this period include the concept of scale-space \citep{Lindeberg1994}, contour models known as snakes \citep{Kass1988}, and projective 3-D reconstructions \citep{Hartley2003}. Researchers also utilized optimization frameworks such as regularization \citep{Tikhonov1977} and Markov random fields \citep{Geman1984}. In addition, statistical learning techniques, like Eigenface \citep{Turk1991}, were employed for facial recognition in images. However, these traditional methods relied on handcrafted feature extraction and shallow models, often struggling to generalize and capture complex patterns in visual data.

The advent of deep learning revolutionized computer vision in recent years, significantly advancing performance and capabilities. Convolutional neural networks (CNNs) \citep{LeCun1998} have enabled automatic learning of hierarchical representations from raw images, bypassing manual feature engineering. This success has been further bolstered by the progress in GPU computing power, which allows for efficient training of increasingly complex and deep models. As a result, deep learning-based computer vision systems have achieved unprecedented success in tasks like object recognition \citep{Krizhevsky2012}, semantic segmentation \citep{Long2015}, and image generation \citep{Radford2015}, surpassing human-level performance in certain benchmarks and enabling practical applications across various sectors.

\begin{figure}[H]
\centering
\caption{Computer Vision vs. Deep Learning}
\includegraphics[width=\linewidth]{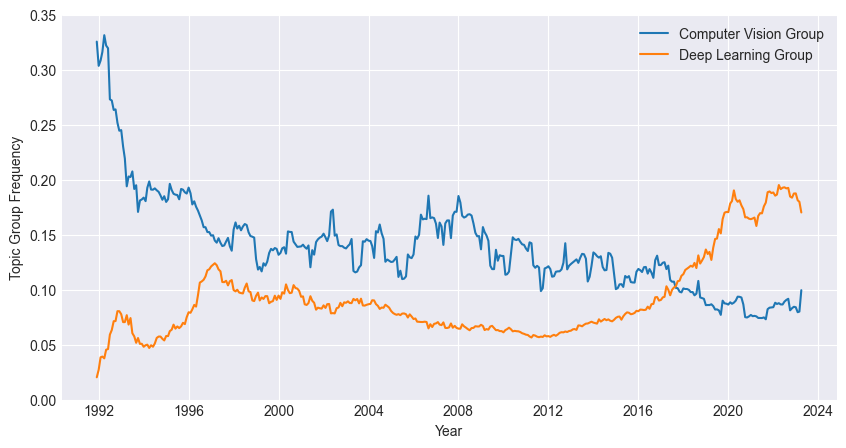}
\label{fig:cv_vs_dl}
\end{figure}

In this study, we manually subdivided topics into ``Groups" or subfields of machine learning, as shown in Tables \ref{table:topics1} and \ref{table:topics2}. The evolution of the \textit{Deep Learning Group} is compared to the \textit{Computer Vision Group} in Figure \ref{fig:cv_vs_dl}. The two series exhibit a statistically significant negative Pearson correlation coefficient of $-0.61749$, suggesting a shift in the scientific community's preference.

During the 1990s, computer vision constituted between a third and a quarter of all publications, while in recent times, it represents merely 10\%. This negative relationship can be attributed to deep learning's capability to automatically learn hierarchical feature representations from raw data, outperforming traditional techniques reliant on manual feature engineering. Consequently, the research focus has shifted toward data-driven methods, leading to a decline in the proportion of theoretical computer vision publications in the field.

\subsubsection{Natural Language Processing}
\label{naturalLanguageProcessing}

The 1950s marked the beginning of NLP as a subfield of artificial intelligence. Alan Turing's test, which involved the automated interpretation and generation of natural language, laid the groundwork for the field \citep{turing50}. At this stage, a fundamental development is the classical and sparse $n$-grams model, which serves as a precursor to the contemporary large language models we are familiar with today \citep{shannon1948mathematical}. Some decades later, we  may cite the research on information retrieval that developed techniques such as TF-IDF (Token Frequency-Inverse Document Frequency)  \citep{Baeza-Yates:2011:MIR:1796408,Manning:2008:IIR:1394399}. 
In the 1980s, NLP shifted towards statistical and machine learning algorithms, driven by increased computational power, the contributions in the field of informational retrieval and the decline of Chomsky's Transformational Grammar linguistic theories \citep{chomsky1965}. 

The 2000s saw a surge in available raw, unannotated language data, prompting a focus on unsupervised and semi-supervised learning algorithms. At this time, different data-driven approaches were applied to deal with important machine learning tasks. Among them, we may cite the matrix factorization based methods  \citep{Gong2001}, the graph based methods \citep{mihalcea2004textrank,Erkan2004}, and the topic modeling based methods \citep{Blei2003}.

Since 2015, NLP has shifted from statistical methods to neural networks, streamlining feature engineering. Techniques such as word embeddings, end-to-end learning of higher-level tasks, and the use of Long Short-Term Memory (LSTM) networks \citep{hochreiter1997long} have gained popularity, leading to significant changes in NLP system design. Deep neural network-based approaches now represent a new paradigm, distinct from statistical natural language processing.

One significant development in NLP is the introduction of attention mechanisms \citep{bahdanau2014neural}, which have improved the performance of models by allowing them to focus on specific parts of input sequences while processing information. The groundbreaking work by \citep{vaswani2017attention}, ``Attention is All You Need", introduced the Transformer architecture, which has revolutionized NLP. Transformers leverage self-attention mechanisms to process input sequences in parallel, rather than sequentially, resulting in improved efficiency and performance, as illustrated in Figure \ref{fig:deep_learning}. Since then, Transformer-based models, such as Google's BERT \citep{devlin2018bert}, OpenAI's GPT-series \citep{radford2018improving, radford2019language}, and numerous other variations, have consistently achieved state-of-the-art results in various NLP tasks, transforming the field and its applications.

\begin{figure}[H]
\centering
\caption{Natural Language Processing Group}
\includegraphics[width=\linewidth]{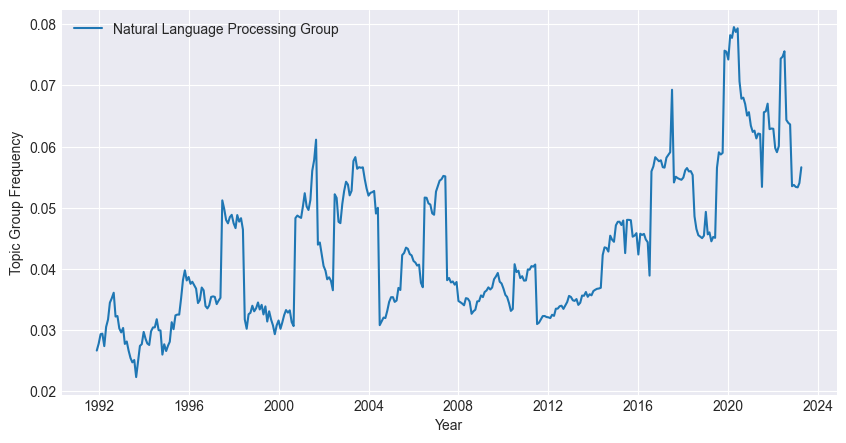}
\label{fig:nlp}
\end{figure}

Figure \ref{fig:nlp} illustrates the evolution of NLP, highlighting that, in contrast to the \textit{Computer Vision Group}, the \textit{Natural Language Processing Group} has experienced a surge in frequency since the 1990s. This is further reinforced by its statistically significant positive Pearson correlation coefficient of $0.7426$ with the \textit{Deep Learning Group}, indicating a strong association between the growth of NLP and the advancements in deep learning techniques.

\subsubsection{Reinforcement Learning}
\label{reinforcementLearning}
Reinforcement Learning (RL) is a subfield of machine learning that focuses on training intelligent agents to make optimal decisions by interacting with their environment. In contrast to supervised learning, which relies on labeled data to learn from, RL is inspired by the trial-and-error learning process observed in humans and animals. The primary components of a reinforcement learning system are an agent, an environment, states, actions, and rewards.

In RL, an agent observes the current state of the environment and takes an action based on its internal policy. The policy, represented by a function, maps states to actions, determining the agent's behavior. After performing an action, the agent receives feedback in the form of a reward signal from the environment. The goal of the agent is to maximize its cumulative reward over time, which requires finding an optimal balance between exploration (trying new actions) and exploitation (relying on actions that have been successful in the past).

The 1990s marked a significant period in the growth of Reinforcement Learning (RL), with groundbreaking advancements shaping the field's trajectory. Q-learning, introduced by Chris Watkins in 1989 \citep{watkins1989learning}, emerged as a key model-free RL algorithm that learns optimal policies without explicitly modeling the environment's dynamics. Furthermore, Richard Sutton's development of Temporal Difference (TD) Learning \citep{sutton1988learning} combined dynamic programming and Monte Carlo methods, allowing agents to learn directly from experience. 

The exploration of function approximation methods, including neural networks, enabled RL algorithms to tackle problems with large state and action spaces. These pivotal developments in the 1990s propelled RL into prominence and solidified its importance in subsequent decades. The 2000s, 2010s, and 2020s witnessed consistent progress in RL, with major breakthroughs such as Deep Q-Networks (DQN) \citep{mnih2015human} and AlphaGo \citep{silver2016mastering} demonstrating the power and versatility of RL algorithms in solving complex real-world problems.

\begin{figure}[H]
\centering
\caption{Reinforcement Learning Group}
\includegraphics[width=\linewidth]{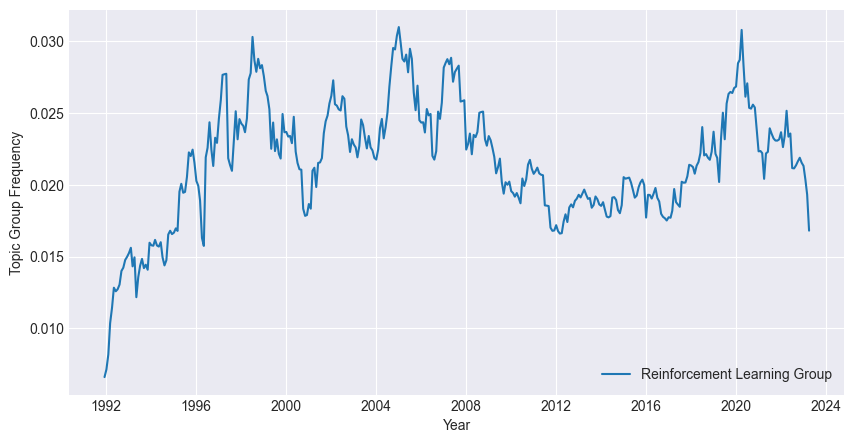}
\label{fig:rl}
\end{figure}

Figure \ref{fig:rl} illustrates the substantial impact of the 1990s' pivotal advancements in Reinforcement Learning (RL) on the field's enduring significance. Notably, the figure reveals a marked surge in RL publication frequency after the transformative innovations of 2015 and 2016, further emphasizing the lasting influence of early RL breakthroughs on the discipline.

\subsubsection{Expert Systems}
\label{expertSystems}

As an early branch of artificial intelligence, Expert Systems emerged in the 1970s and 1980s, focusing on developing rule-based systems emulating human expert decision-making capabilities \citep{hayesroth1983building, jackson1986principles}. These systems, comprising a knowledge base, an inference engine, and a user interface, captured domain-specific knowledge in rules and facts, applying logical reasoning to draw inferences and provide recommendations. While expert systems played a relevant role in AI's historical evolution, their influence in the current machine learning landscape has diminished due to the advent of more sophisticated techniques like representation learning \citep{lecun2015deep}.

Representation learning is a method in which models automatically discover and learn relevant features or representations from raw data, without relying on manually engineered features. This contrasts with the feature engineering approach used in Expert Systems, where domain experts would design and handcraft features to capture the most relevant aspects of the problem. The shift towards representation learning has allowed for more flexible, scalable, and adaptive models capable of handling complex, high-dimensional data.

\begin{figure}[H]
\centering
\caption{Expert Systems \& Design Topic}
\includegraphics[width=\linewidth]{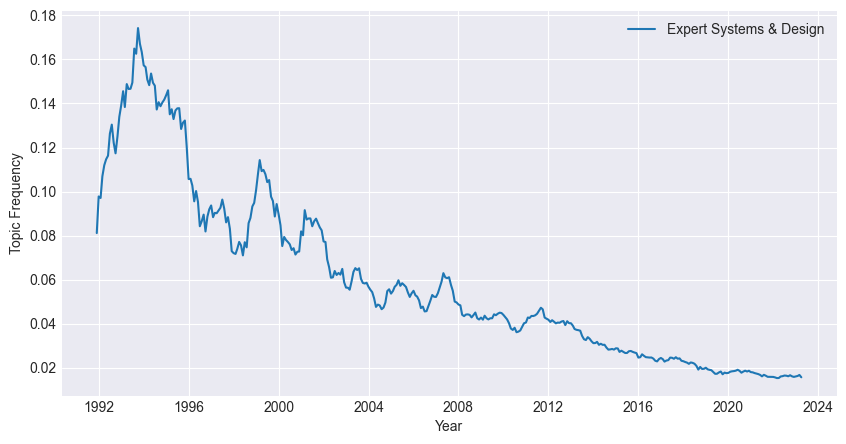}
\label{fig:ESDesign}
\end{figure}

In the 1990s, expert systems represented one of the most frequent topics in machine learning literature, peaking at 17.42\% of all publications. During this period, expert systems gained widespread recognition and were employed in various applications, such as medical diagnosis \citep{miller1987internist}, business decision-making \citep{wong1995expert}, and fault detection \citep{venkatasubramanian2003review}. However, today, they account for only around 1.53\%, as shown in Figure \ref{fig:ESDesign}, highlighting the decline in this topic's importance. The shift in focus towards representation learning and the rise of deep learning have contributed to the reduced emphasis on expert systems in contemporary AI research.

\subsubsection{Historical Contexts}
\label{historicalContexts}

Examining the evolution of scientific ideas, methodologies, and paradigms provides researchers with insights into factors influencing past discoveries, limitations of prevailing techniques, and driving forces behind major shifts in scientific thinking. Understanding historical context allows scientists to appreciate current theories and practices, identify foundations for new knowledge, and recognize the social, economic, and political forces shaping scientific inquiry. This contextual awareness deepens the understanding of the scientific process, informs future research direction, and fosters a holistic and nuanced perspective on scientific progress.

\begin{figure}[H]
\centering
\caption{Financial Markets \& Risk Topic}
\includegraphics[width=\linewidth]{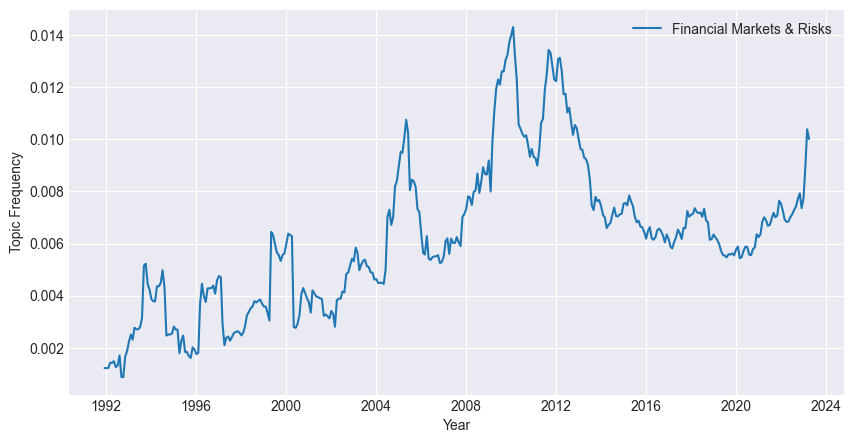}
\label{fig:fin_markt}
\end{figure}

As Figure \ref{fig:fin_markt} demonstrates, the 2008 financial crisis precipitated a marked increase in machine learning publications addressing financial markets and risk. This event exposed the inadequacies of conventional risk management and forecasting methods, prompting researchers to seek innovative solutions. Machine learning proved to be a powerful resource, offering precise, data-driven insights for market trends and decision-making processes, which exemplifies how historical events can significantly influence the trajectory of scientific research and technology within a particular domain.

\begin{figure}[H]
\centering
\caption{Medical Diagnosis and Patient Health}
\includegraphics[width=\linewidth]{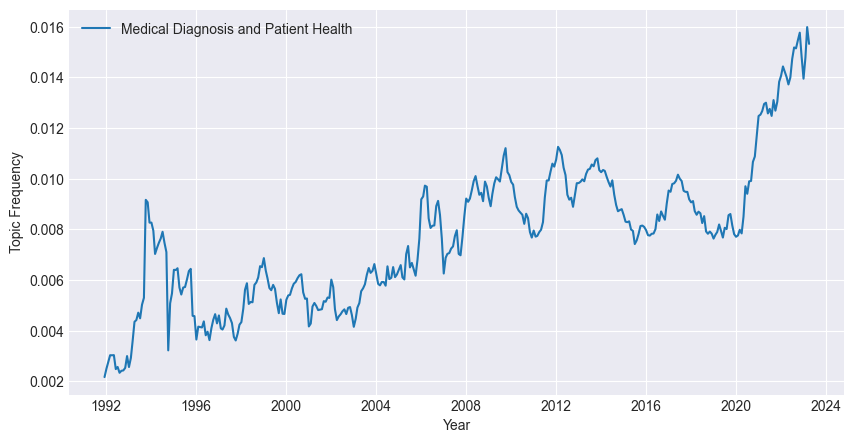}
\label{fig:med_diagnosis}
\end{figure}

Figure \ref{fig:med_diagnosis} highlights the substantial increase in machine learning publications focused on medical diagnosis and patient health following the 2020 COVID-19 pandemic. This event emphasized the necessity for advanced diagnostic tools and personalized healthcare solutions, leading researchers to explore machine learning applications in disease detection, treatment, and patient care \citep{vaishya2020artificial, alimadadi2020artificial,khan2021applications,utku2023deep}. This example further illustrates the profound influence of historical context on scientific progress and technology development within specific fields, such as healthcare and medical diagnosis.

\subsection{Novelty, Transience, Resonance}
\label{results-ntr}

\begin{figure}[H]
    \centering
    \caption{Innovation Bias for $w=12$}
    \begin{subfigure}[b]{0.45\linewidth}
        \label{innovation_bias_w=12_a}
        \centering
        \includegraphics[width=\linewidth]{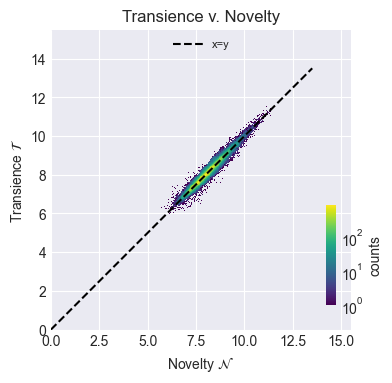}
        \caption{}
        \label{fig:innovation_bias_a}
    \end{subfigure}
    \hfill
    \begin{subfigure}[b]{0.45\linewidth}
        \label{innovation_bias_w=12_b}
        \centering
        \includegraphics[width=\linewidth]{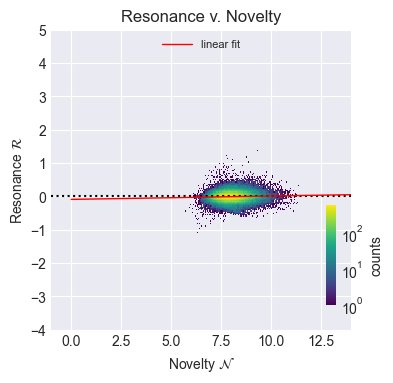}
        \caption{}
        \label{fig:innovation_bias_b}
    \end{subfigure}
    \label{fig:innovation_bias}
\end{figure}

Figure \ref{fig:innovation_bias_a} shows that the relation between Transience and Novelty is close to the identity line ($x=y$).   This suggests that an increase in novelty is generally matched by an equal increase in transience. In simpler terms, the more novel a research work is, the less likely it is for that content to propagate into subsequent works. However, this symmetry is broken by resonant works, which differ more from their past and align more with their future. These works are found below the identity line, where novelty outweighs transience.

In Figure \ref{fig:innovation_bias_b}, we see that the red line, representing the  novelty effectiveness defined in Equation \ref{eq:novelty_effectiveness}, is close to zero. This indicates that there is no systematic relation between novelty and resonance in the entire dataset. Despite the general trend of increased novelty leading to increased transience, the lack of a systematic relationship between novelty and resonance suggests that the influence of a paper is not solely determined by its novelty.

\subsubsection{Authors}

In order to evaluate the author capabilities, we need to attribute publications to their respective authors. However, the extensive diversity of venues in the dataset presented challenges in accurately matching authors, particularly when dealing with identical names or those publishing under multiple name variations (e.g., YOSHUA, BENGIO and BENGIO, YOSHUA). With 225,825 unique names in the dataset, manually verifying each case was not feasible. The difficulty of dealing with unmatching names is a well-known issue in scientific research involving large datasets \citep{tang2010, strotmann2012}. To address this challenge, several measures were taken to improve the accuracy of author name matching.

Initially, the focus was narrowed to the top 1000 authors with the largest number of publications. Due to some authors having the same number of papers, this reduced dataset comprised 1039 unique author names. Subsequently, the \cite{dblp} API and \cite{athenianco_names_matcher}'s Names Matching Fuzzy Algorithm\footnote{This algorithm is structured around the subsequent steps: (1) Parsing, normalizing, and segmenting the names within each identity, resulting in a set of strings for each one. (2) Establishing the similarity between identities. (3) Creating the distance matrix between identities within two designated lists. (4) Addressing the Linear Assignment Problem (LAP) associated with this matrix.} were employed to automatically identify duplicate names, further reducing the unique names to 1029.

\begin{figure}[H]
\centering
\caption{Innovation Bias for the 1029 most frequent authors}
\includegraphics[width=0.7\linewidth]{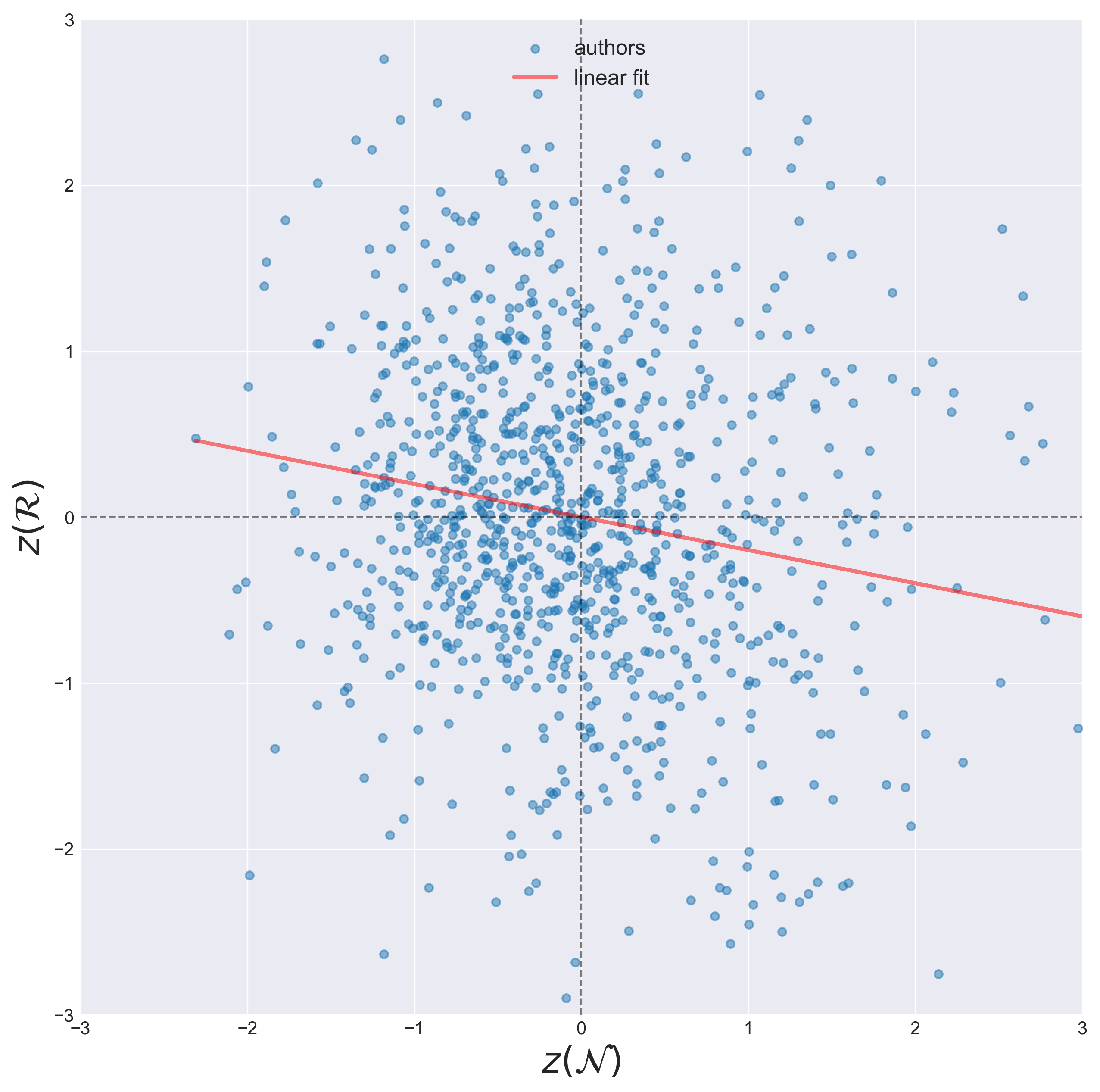}
\label{fig:authors_bias}
\end{figure}

It is important to emphasize that the comparison being conducted pertains to the most prolific researchers in the field, whose contributions undoubtedly hold considerable significance. Nonetheless, within this group of scholars,  Figure \ref{fig:authors_bias} reveals a notable observation: among the 1029 most frequent writers, there appears to be a significant novelty avoidance. This result suggests that established authors may have a preference for working within their areas of expertise and familiarity, leading to a more conservative approach in their research and a lower degree of novelty compared to less-established researchers who are more likely to explore uncharted territory or take risks with novel ideas. 

Additionally, the top 1000 authors, who may have a higher degree of influence in their respective fields, could be more focused on refining and consolidating existing knowledge rather than pursuing radical innovations, potentially stemming from the pressure to maintain their status and reputation within the scientific community. However, it is important to note that some authors defied this tendency and managed to achieve high resonance in their work, even as they pursued high novelty.

\begin{table}
\centering
\begin{adjustbox}{width=\textwidth}
\begin{minipage}{\textwidth}

\caption{Highest and lowest scoring authors for Novelty and Resonance}
\label{table:ntr_authors}
\begin{tabular}{l|llll|llll|}
                              & \multicolumn{4}{c}{High resonance}                                                & \multicolumn{4}{c}{Low resonance}                                               \\
                              \midrule
                              & Name              & $z(\mathcal{N})$ & $z(\mathcal{R})$ & $\Delta z(\mathcal{N})$ & Name            & $z(\mathcal{N})$ & $z(\mathcal{R})$ & $\Delta z(\mathcal{N})$
                              \\
                              \midrule
\multirow{5}{*}{\rotatebox{90}{High novelty}} & Qiu, Xipeng     & 1.352 & 2.394 & 2.664 & Tong, Shaocheng          & 2.138 & -2.755 & -2.328 \\
                              & Huang, Xuanjing    & 1.299 & 2.268 & 2.527 & Chen, Huayou    & 6.271 & -2.572 & -1.321 \\
                              & Sun, Xu & 1.256 & 2.102 & 2.353 & Hua, Changchun    & 1.201 & -2.500 & -2.260 \\
                              & Zhao, Dongyan        & 1.796 & 2.026 & 2.385 & Liao, Huchang  & 5.121 & -2.496 & -1.474 \\
                              & Wang, William Yang          & 1.490 & 1.999 & 2.297 & Mesiar, Radko     & 4.408 & -2.439 & -1.560 \\
                              \midrule
                              & Name              & $z(\mathcal{N})$ & $z(\mathcal{R})$ & $\Delta z(\mathcal{N})$ & Name            & $z(\mathcal{N})$ & $z(\mathcal{R})$ & $\Delta z(\mathcal{N})$ \\
                              \midrule
\multirow{5}{*}{\rotatebox{90}{Low novelty}}  & Pang, Yanwei     & -1.181 & 2.760 & 2.524 & Hsu, Chun-fei   & -1.182 & -2.634 & -2.870 \\
                              & Lu, Huchuan  & -1.085 & 2.394 & 2.178 & Raja, Muhammad A. Z.   & -1.986 & -2.158 & -2.555 \\
                              & Ouyang, Wanli  & -1.350 & 2.272 & 2.002 & Veeraraghavan, Ashok   & -1.146 & -1.917 & -2.145 \\
                              & Shen, Jianbing & -1.255 & 2.214 & 1.964 & Li, Kenli   & -1.301 & -1.571 & -1.831 \\
                              & Huang, Feiyue      & -1.580 & 2.013 & 1.697 & Hu, Bin   & -1.834 & -1.394 & -1.760 \\
                              \bottomrule
\end{tabular}%
\end{minipage}
\end{adjustbox}

\end{table}

Table \ref{table:ntr_authors} displays the authors with the highest and lowest scores in terms of novelty and resonance. We identified these authors by applying a z-score transformation to the data, selecting the top 100 most novel authors and the top 100 least novel authors. We then chose the 5 authors with the highest resonance and the 5 authors with the lowest resonance from each group.

The metric $\Delta z(\mathcal{N})$, defined as $z(\mathcal{R}) - E[z(\mathcal{R})|z(\mathcal{N})]$, quantifies the deviation of an author's resonance score from the expected resonance score given their novelty score. Essentially, it shows the extent to which an author's resonance diverges from the overall trend observed between novelty and resonance in the data.

Observing a high $\Delta z(\mathcal{N})$ for authors with high novelty implies that these authors are able to achieve a greater than expected impact on their field, even as they pursue novel ideas. This deviation from the general trend of innovation avoidance might be attributed to the individual abilities and skills of these authors, which enable them to explore new concepts while still making a significant impact in their respective domains.

\subsubsection{Venues}

\begin{figure}[H]
\centering
\caption{Innovation Bias for venues}
\includegraphics[width=0.7\linewidth]{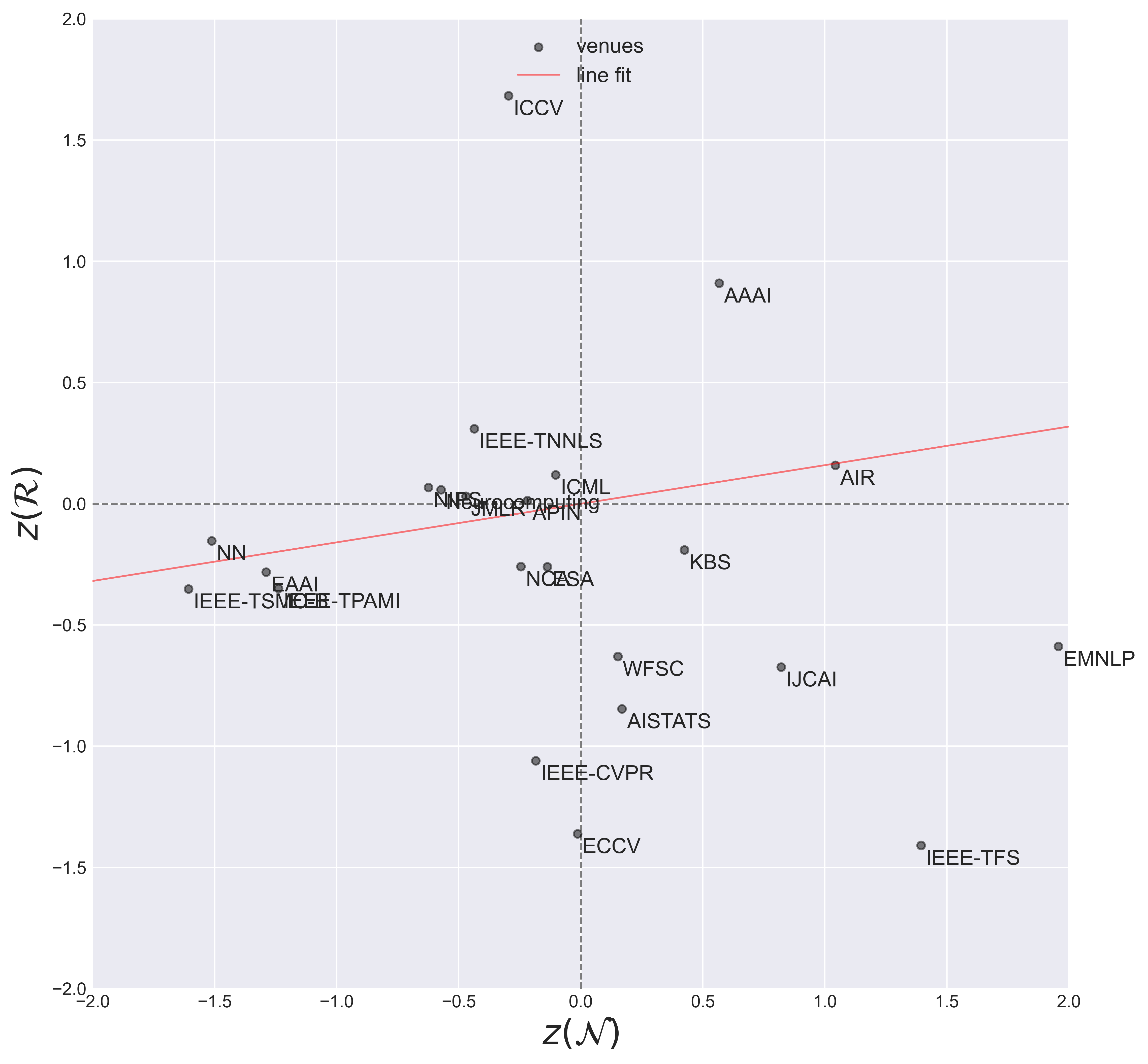}
\label{fig:venues_bias}
\end{figure}

To be able to compare the academic venues in terms of resonance and novelty, we evaluate these metrics for individual papers using the information of the topic of the paper. Subsequently, we categorize the papers based on their respective venues. Finally, we calculate the average resonance and novelty values for each specific venue. These values are presented in Figure \ref{fig:venues_bias}, along with the linear regression of these values. This regression can provide valuable insights into whether a given venue falls above or below the common average. It is important to acknowledge the presence of inherent sampling bias in our analysis. The selected venues were included based on their perceived significance, influence, and anticipated novelty in the field of Machine Learning. Therefore, in this figure, we are inherently comparing highly relevant venues. Deviations from the straight line should not be interpreted as diminishing a venue's worthiness, given the distinguished nature of the venues being compared.

 Furthermore, it is worth mentioning the unique nature of the Computer Science field, which, unlike many other fields, accords conferences a distinctive significance. Notably, numerous significant findings are exclusively disseminated through conferences in this field due to their rapid information dissemination.

Table \ref{table:ntr_venues_ranking} presents the venues considered in this work ranked by the deviations to the expected resonance score given their novelty score. It is amazing the relevant role of conferences in specific fields, such as ICLR  (International   Conference on Learning Representations), ACL (Meeting of the Association for Computational Linguistics) ICCV (International Conference on Computer Vision). 

\begin{sidewaystable*}[]
\centering
\caption{Venues normalized novelties and resonances ranked by $\Delta   z(\mathcal{R})$}
\label{table:ntr_venues_ranking}
\resizebox{\linewidth}{!}{%
\begin{tabular}{@{}l|l|l|l|l@{}}
Venue                                                                      & Abbreviation   & $z(\mathcal{N})$ & $z(\mathcal{R})$ & $\Delta   z(\mathcal{R})$ \\
\toprule

International Conference on Learning   Representations                     & ICLR           & -0.12159         & 2.587853         & 2.607214                \\
Meeting of the Association for   Computational Linguistics (ACL)           & ACL            & 2.539549         & 2.471487         & 2.067111                \\
International Conference on Computer   Vision (ICCV)                       & ICCV           & -0.29588         & 1.682636         & 1.729749                \\
AAAI Conference on Artificial   Intelligence                               & AAAI           & 0.567189         & 0.910322         & 0.820008                \\
IEEE Transactions on Neural Networks and   Learning Systems                & IEEE-TNNLS     & -0.43608         & 0.309849         & 0.379287                \\
Neural Information Processing Systems                                      & NIPS           & -0.62429         & 0.06708          & 0.166486                \\
Neurocomputing                                                             & Neurocomputing & -0.57209         & 0.058609         & 0.149704                \\
International Conference on Machine   Learning                             & ICML           & -0.10303         & 0.119577         & 0.135983                \\
Journal of Machine Learning Research                                       & JMLR           & -0.47041         & 0.031591         & 0.106495                \\
Neural Networks                                                            & NN             & -1.51144         & -0.15239         & 0.088282                \\
Applied Intelligence                                                       & APIN           & -0.21868         & 0.013124         & 0.047944                \\
Artificial Intelligence Review                                             & AIR            & 1.042796         & 0.158838         & -0.00721                \\
Engineering Applications of Artificial   Intelligence                      & EAAI           & -1.28932         & -0.28147         & -0.07617                \\
IEEE Transactions on Systems, Man, and   Cybernetics, Part B (Cybernetics) & IEEE-TSMC-B    & -1.6068          & -0.35138         & -0.09553                \\
IEEE Transactions on Pattern Analysis and   Machine Intelligence           & IEEE-TPAMI     & -1.2382          & -0.34967         & -0.15251                \\
Neural Computing and Applications                                          & NCA            & -0.24493         & -0.25899         & -0.21999                \\
Expert Systems with Applications                                           & ESA            & -0.13667         & -0.25988         & -0.23812                \\
Knowledge-Based Systems                                                    & KBS            & 0.42484          & -0.19057         & -0.25822                \\
Applied Soft Computing                                                     & WFSC           & 0.15193          & -0.62989         & -0.65409                \\
International Joint Conferences on   Artificial Intelligence Organization  & IJCAI          & 0.821203         & -0.67314         & -0.8039                 \\
International Conference on Artificial   Intelligence and Statistics       & AISTATS        & 0.168854         & -0.8456          & -0.87248                \\
Conference on Empirical Methods in   Natural Language Processing           & EMNLP          & 1.95683          & -0.58854         & -0.90013                \\
Computer Vision and Pattern Recognition   Conference                       & IEEE-CVPR      & -0.18398         & -1.05983         & -1.03053                \\
European Conference on Computer Vision                                     & ECCV           & -0.01344         & -1.36115         & -1.35901                \\
IEEE Transactions on Fuzzy Systems                                         & IEEE-TFS       & 1.393643         & -1.40848         & -1.63039               \\

\bottomrule
\end{tabular}
}
\end{sidewaystable*}

\section{Conclusions}
\label{sec:conclusion}

In this study, we have employed LDA to delve into the evolution of machine learning (ML) research. Through LDA, we discern key themes and foundational concepts within the field. By segmenting these themes, we trace their temporal trends. Ultimately, leveraging the Kullback-Leibler Divergence metric, we ascertain the roles of prominent authors and machine learning venues in shaping the landscape.

Our findings unveil the swift evolution of the machine learning field towards emerging technologies, occurring concurrently with the diminishing relevance of other technologies that are gradually receding from the forefront. Remarkably, deep learning emerges as the focal point of utmost interest within the field, while expert systems, once of paramount importance, have irreversibly slipped into obscurity. Moreover, domains such as computer vision and natural language processing have substantially integrated into the realm of deep learning research.

We have also investigated the roles of the authors in generating novel insights and the academic venues for disseminating this knowledge. Notably, our exploration has revealed that certain prominent authors exhibit an inclination towards  innovation, while others adopt a more conventional stance. Regarding academic venues, we have identified the distinct significance of conferences and broadly scoped periodicals in spreading this wealth of knowledge.

 \section*{Acknowledgements}
D.O.C. is indebt to CNPQ for partial financial support (302629/2019-0).

\clearpage

\begin{appendices}

\counterwithin{figure}{section}
\counterwithin{table}{section}
\counterwithin{equation}{section}

\label{Appendix}

\section{List of popular machine leaning sources}
\label{app:urls}

Table \ref{table:urls} presents a list of the popular machine learning sources used to find the list of most popular machine learning venues.

\begin{table}
\begin{adjustbox}{width=\textwidth}
\begin{minipage}{\textwidth}
\caption{URL's used in the first step of our search}
\label{table:urls}
\begin{tabular}{|l|}
\hline
\url{https://aclanthology.org/} \\
\hline
\url{https://proceedings.mlr.press/} \\
\hline
\url{https://www.quora.com/What-are-the-best-conferences-and-journals-about-machine-learning} \\
\hline
\url{https://research.com/conference-rankings/computer-science/machine-learning} \\
\hline
\url{https://deepai.space/top-ai-conferences-and-journals/} \\
\hline
\url{https://www.junglelightspeed.com/the-top-10-nlp-conferences/} \\
\hline
\end{tabular}%
\end{minipage}
\end{adjustbox}

\end{table}

\section{Details of the LDA implementation}
\label{app:ldaDetails}

In this section, we describe the process we used to determine the values of the LDA parameters. 

We start by applying Document Frequency (TF) and Token Frequency-Inverse Document Frequency (TF-IDF) in order to eliminate words exhibiting low importance within the corpus. Then, we search for the optimal value of the number of topics $K$. 

As we mention before, we have implemented the LDA using the Python library Gensim. It
provides three alternatives for setting priors: (1) `symmetric' (default), utilizing a fixed symmetric prior of ${1}/{num\_topics}$, (2) `asymmetric', implementing a fixed normalized asymmetric prior of ${1}/{(topic\_index + \sqrt{num\_topics})}$, and (3) `auto', which learns an asymmetric prior from the corpus. We incorporate these these configurations alongside document frequency and tf-idf filters to optimize the LDA model's performance during the hyperparameter search process.

Table \ref{table:hyperparameters} displays the results, with document frequency values set at 0.5, meaning words appearing in more than 50\% of the documents were removed from the corpus. The second value, 1.0, indicates no words were removed for models trained with this parameter. The tf-idf parameter compared a low value (0.0075), which removed fewer words, against a high value (0.015), which removed more words. The search space for $\alpha$ was (`symmetric', `asymmetric') and for $\eta$ it was (`symmetric', `auto'). The best-performing model had $K=60$, $\text{df}=0.5$, $\text{tf-idf}=0.0075$, $\alpha=\text{asymmetric}$, and $\eta=\text{auto}$. The reader can refer to Tables \ref{table:topics1} and \ref{table:topics2} to examine the resulting topics derived from this optimal combination of parameters, please access this paper's \href{https://zenodo.org/record/8298911}{Zenodo} for a more detailed overview on the topics and their word distributions.

\begin{table}
\centering
\caption{Parametric space search results}
\label{table:hyperparameters}
\begin{tabular}{@{}c|c|c|c|c|c@{}}
\toprule
DF  & TF-IDF & $K$ & $\alpha$   & $\eta$    & $C_V$ coherence \\
\midrule
0.5 & 0.0075 & 60  & symmetric  & symmetric & 0.531903                         \\
0.5 & 0.0075 & 60  & symmetric  & auto      & 0.531045                         \\
0.5 & 0.0075 & 60  & asymmetric & symmetric & 0.545108                         \\
\textbf{0.5} & \textbf{0.0075} & \textbf{60}  & \textbf{asymmetric} & \textbf{auto}      & \textbf{0.552183}                        \\
0.5 & 0.015  & 60  & symmetric  & symmetric & 0.524149                         \\
0.5 & 0.015  & 60  & symmetric  & auto      & 0.543684                         \\
0.5 & 0.015  & 60  & asymmetric & symmetric & 0.528428                         \\
0.5 & 0.015  & 60  & asymmetric & auto      & 0.523044                         \\
1   & 0.0075 & 60  & symmetric  & symmetric & 0.526615                         \\
1   & 0.0075 & 60  & symmetric  & auto      & 0.536799                         \\
1   & 0.0075 & 60  & asymmetric & symmetric & 0.527374                         \\
1   & 0.0075 & 60  & asymmetric & auto      & 0.520164                         \\
1   & 0.015  & 60  & symmetric  & symmetric & 0.532143                         \\
1   & 0.015  & 60  & symmetric  & auto      & 0.534907                         \\
1   & 0.015  & 60  & asymmetric & symmetric & 0.518847                         \\
1   & 0.015  & 60  & asymmetric & auto      & 0.504365   
                    \\
\bottomrule
\end{tabular}

\end{table}

\begin{figure}[H]
  \centering
  \caption{Model Coherence vs. K - Number of Topics}
  \includegraphics[width=\linewidth]{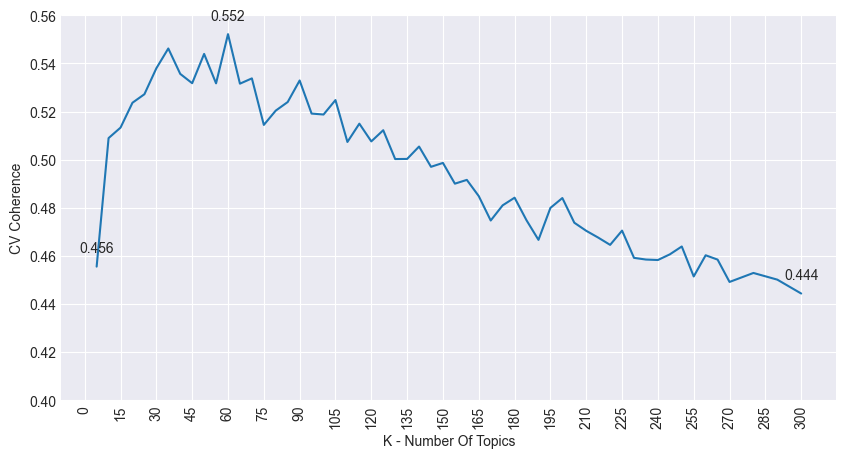}
  \label{fig:coherence_K}
\end{figure}

Figure \ref{fig:coherence_K} displays graphically the results of the search for $K$ values ranging from 5 to 300, with coherence peaking at $0.552$ for $K=60$. Lower $K$ values fail to adequately capture the complex structure of the machine learning field's literature, likely combining disparate topics. As the $K$ value increases, the optimal representation of the underlying semantic structure diminishes, resulting in topics primarily driven by statistical co-occurrence.

\section{Summary of topics}
\label{venues_individual}

Tables \ref{table:topics1} and \ref{table:topics2} display the topic labels generated by a collaboration between researchers and ChatGPT. The researchers provided the most frequent words for each topic, and ChatGPT, leveraging its language understanding capabilities, responded with an appropriate label for the topic. After that, the authors reviewed the labels generated by ChatGPT. These tables offer a succinct and interconnected representation of the dataset, showcasing the associated groups and the top words within each topic, thus providing a more comprehensive overview of the underlying themes in the data. Groups with an * were  not decisively classified.

\begin{sidewaystable*}[]
\centering
\caption{Summary of topics, groups and the main terms in each topic.}
\label{table:topics1}
\resizebox{\linewidth}{!}{%
\begin{tabular}{@{}r|l|l|l@{}}
Topics & Labels                                             & Groups             & Terms                                                \\
\toprule
0      & Graph Theory and Network Structures                & Networks                    & graph, structure, node, tree, edge                   \\
1      & Control Systems and Dynamic Systems                & Systems \& Control          & control, system, time, controller, design            \\
2      & Systems and Energy Management                      & Systems \& Control          & system, power, time, energy, sensor                  \\
3      & Multi-level Feature Fusion and Deep Learning       & Deep Learning               & multi, level, feature, module, dataset               \\
4      & Probability Distribution and Estimation Algorithms & Statistics                  & distribution, algorithm, probability, show, estimate \\
5      & Evolutionary Algorithm and Optimization            & Genetic Algorithms          & algorithm, optimization, search, performance, local  \\
6      & Urban Navigation and Localization                  & Navigation                  & location, trajectory, vehicle, mobile, road          \\
7      & Object Recognition and Visual Saliency             & Computer Vision             & object, visual, image, scene, category               \\
8      & Object Detection                                   & Computer Vision             & detection, detect, detector, community, proposal     \\
9      & Cybersecurity \& Network Communication             & Cybersecurity               & event, attack, service, communication, distribute    \\
10     & Information Retrieval \& Data Mining               & Data*                       & pattern, query, search, retrieval, question          \\
11     & Image Processing and Resolution                    & Computer Vision             & surface, resolution, image, color, light             \\
12     & Relational Data and Entity Relationships           & Data*                       & group, relation, entity, link, knowledge             \\
13     & Failure Diagnosis and Causal Effects               & Fault Diagnosis             & fault, effect, variable, failure, causal             \\
14     & Medical Diagnosis and Patient Health               & Medical                     & diagnosis, patient, medical, disease, clinical       \\
15     & Neural Network and Synaptic Functioning            & Deep Learning               & neuron, spike, mechanism, unit, channel              \\
16     & Time Series Forecasting and Prediction Modeling    & Statistics                  & prediction, time, series, model, predict             \\
17     & Camera Tracking and Motion Estimation              & Computer Vision             & camera, tracking, point, depth, image                \\
18     & Bias and Evidence Analysis                         & Statistics                  & bias, evidence, hypothesis, explanation, argument    \\
19     & Game Theory and Strategic Behavior                 & Decision Making             & strategy, game, play, player, equilibrium            \\
20     & Facial Recognition \& Identification               & Computer Vision             & face, recognition, image, identification, person     \\
21     & Computational Efficiency \& Performance            & Complexity                  & time, large, computational, scale, reduce            \\
22     & Shape and Gesture Recognition                      & Computer Vision             & shape, recognition, hand, body, line                 \\
23     & Semantic Word Embeddings in NLP                    & Natural Language Processing & word, semantic, sentence, embedding, representation  \\
24     & Reinforcement Learning and Robotics                & Reinforcement Learning      & agent, policy, environment, learning, robot          \\
25     & Human Attributes \& Interactions                   & Reinforcement Learning*     & human, attribute, cell, interaction, behavior        \\
26     & Neural Networks                                    & Deep Learning               & network, neural, layer, architecture, deep           \\
27     & Expert Systems \& Design                           & Expert Systems              & system, knowledge, design, process, expert           \\
28     & Regression Analysis \& Modeling                    & Statistics                  & error, regression, model, test, parameter            \\
29     & Information Imputation and Completion              & Machine Learning            & information, miss, mutual, incomplete, side         \\
\bottomrule
\end{tabular}
}
\end{sidewaystable*}

\begin{sidewaystable*}[]
\centering
\caption{Summary of topics, fields and the main terms in each topic.}
\label{table:topics2}
\resizebox{\linewidth}{!}{%
\begin{tabular}{@{}r|l|l|l@{}}
Topics & Labels                                 & Groups             & Terms                                                      \\
\toprule
30     & Constraint Planning \& Logic           & Logic                       & constraint, path, show, set, property                      \\
31     & Feature Extraction \& Selection        & Machine Learning            & feature, extract, selection, extraction, select            \\
32     & Adversarial Generative Modeling        & Deep Learning               & generate, model, generation, adversarial, training         \\
33     & Sentiment and Emotion Analysis         & Natural Language Processing & model, dataset, sentiment, emotion, task                   \\
34     & Probabilistic Modeling \& Inference    & Statistics                  & model, parameter, process, modeling, inference             \\
35     & Social Recommendation Systems          & Recommendation Systems      & user, recommendation, item, social, system                 \\
36     & Classification \& Performance          & Machine Learning            & classification, classifier, machine, accuracy, performance \\
37     & Multi-Dimensional Data Representation  & Statistics                  & space, representation, view, dimensional, sparse           \\
38     & Robust Denoising                       & Statistics                  & noise, loss, robust, noisy, robustness                     \\
39     & Data Clustering \& Segmentation        & Machine Learning            & cluster, clustering, algorithm, datum, partition           \\
40     & Image Partitioning \& Texture Analysis & Computer Vision             & image, segmentation, region, pixel, texture                \\
41     & Decision Making \& Evaluation          & Decision Making             & decision, criterion, make, preference, selection           \\
42     & Speech \& Language Processing          & Natural Language Processing & language, code, translation, model, speech                 \\
43     & Domain Adaptation \& Transfer Learning & Machine Learning*           & domain, target, source, transfer, adaptation               \\
44     & Semi-Supervised Data Annotation        & Machine Learning            & label, datum, training, semi\_supervised, annotation       \\
45     & Modeling with Attention Mechanisms     & Deep Learning               & sequence, attention, model, long, memory                   \\
46     & Deep Learning                          & Deep Learning               & learning, learn, task, training, deep                      \\
47     & Signal Processing                      & Synaptic Functioning*       & signal, phase, brain, frequency, activity                  \\
48     & Video Action Recognition               & Computer Vision             & video, temporal, action, frame, motion                     \\
49     & Document Summarization                 & Natural Language Processing & document, topic, content, web, summary                     \\
50     & Similarity Measures and Rankings       & Statistics*                 & measure, similarity, metric, correlation, rank             \\
51     & Research and Development Trends        & Research                    & research, study, analysis, application, technique          \\
52     & Fuzzy Logic Systems                    & Fuzzy Algorithms            & fuzzy, rule, system, function, logic                       \\
53     & Fuzzy Value \& Set Theory              & Fuzzy Algorithms            & value, set, weight, interval, uncertainty                  \\
54     & Linear and Nonlinear Optimization      & Optimization                & function, matrix, linear, problem, non                     \\
55     & Financial Markets \& Risks             & Finance                     & risk, market, student, financial, price                    \\
56     & Adaptive optimization algorithms       & Optimization                & algorithm, parameter, rate, filter, convergence            \\
57     & Class Imbalance \& Sampling            & Machine Learning            & class, sample, instance, problem, learning                 \\
58     & Data Management and Utilization        & Data                        & datum, data, set, real, technique                          \\
59     & Optimization \& Problem-Solving        & Optimization                & problem, algorithm, solution, solve, optimization         \\
\bottomrule
\end{tabular}
}
\end{sidewaystable*}

\end{appendices}

\clearpage


\section*{Declarations}

\begin{itemize}
\item Funding: DOC thanks to cnpq for financial support (302629/2019-0).
\item Conflict of interest/Competing interests: The authors declare that they have no conflict of interest. 
\item Ethics approval: The manuscript has not been simultaneously submitted or published in any other journal.
\item Consent to participate: ``Not applicable''.
\item Consent for publication: All authors have read and have approved the manuscript being submitted and agree to its submittal to this journal.
\item Availability of data and materials: We may find the complete dataset at  \href{https://zenodo.org/record/8298911}{Zenodo}. 
\item Code availability We may find the complete code at  \href{https://zenodo.org/record/8298911}{Zenodo}.
\item Authors' contributions:  

\begin{itemize}
    \item Conceptualization: [V\'itor Bandeira Borges, Daniel Oliveira Cajueiro]; 
    \item Methodology: [V\'itor Bandeira Borges, Daniel Oliveira Cajueiro]; 
    \item Formal analysis and investigation: [V\'itor Bandeira Borges];
    \item Writing - original draft preparation: [V\'itor Bandeira Borges];
    \item Writing - review and editing: [V\'itor Bandeira Borges, Daniel Oliveira Cajueiro]; 
    \item Supervision: [Daniel Oliveira Cajueiro].
\end{itemize}

\end{itemize}

\noindent

\bibliography{sn-bibliography}

\end{document}